\documentclass[conference]{IEEEtran}
\IEEEoverridecommandlockouts
\usepackage{cite}
\usepackage{amsmath,amssymb,amsfonts}
\usepackage{graphicx}
\usepackage{textcomp}
\usepackage{xcolor}
\usepackage{booktabs}
\usepackage{multirow}
\usepackage{paralist}
\usepackage{caption}
\usepackage{setspace}
\usepackage[ruled,vlined,linesnumbered]{algorithm2e}
\usepackage{relsize}
\usepackage{marvosym}
\usepackage{amssymb}

\usepackage{hyperref}

\def\BibTeX{{\rm B\kern-.05em{\sc i\kern-.025em b}\kern-.08em
    T\kern-.1667em\lower.7ex\hbox{E}\kern-.125emX}}
\usepackage{xspace}
\usepackage{color}
\usepackage{multirow}
\usepackage{ifthen}
\usepackage{paralist}
\renewcommand{\baselinestretch}{1}

\long\def\ignorethis#1{}

\definecolor{gray}{rgb}{0.35,0.35,0.35}
\definecolor{MyBlue}{rgb}{0,0.2,0.8}
\definecolor{MyRed}{rgb}{0.8,0.2,0}
\definecolor{MyGreen}{rgb}{0.0,0.5,0.1}
\definecolor{MyGray}{rgb}{0.4,0.4,0.4}
\definecolor{purple}{RGB}{112,48,160}

\newlength\paramargin
\newlength\figmargin
\newlength\subfigmargin
\newlength\secmargin
\newlength\subsecmargin
\newlength\tabmargin
\newlength\eqmargin

\usepackage{array}
\newcolumntype{L}[1]{>{\raggedright\let\newline\\\arraybackslash\hspace{0pt}}m{#1}}
\newcolumntype{C}[1]{>{\centering\let\newline\\\arraybackslash\hspace{0pt}}m{#1}}
\newcolumntype{R}[1]{>{\raggedleft\let\newline\\\arraybackslash\hspace{0pt}}m{#1}}

\newcommand{\subsecref}[1]{Section~\ref{subsec:#1}}
\newcommand{\figref}[1]{Fig.~\ref{fig:#1}}
\newcommand{\tabref}[1]{Table~\ref{tab:#1}}
\newcommand{\eqnref}[1]{Eq.\eqref{eq:#1}}

\newcommand{\algref}[1]{Algorithm~\ref{#1}}

\newcommand{\Paragraph}[1]{\noindent\textbf{#1}}

\definecolor{mycolor_blue}{RGB}{231,239,250}
\definecolor{mycolor_green}{RGB}{230,247,224}
\definecolor{mycolor_gray}{RGB}{236,236,236}
\definecolor{pearDark!20}{RGB}{212,230,241}

\begin{document}

\title{MLV-Edit: Towards Consistent and Highly Efficient Editing for Minute-Level Videos}

\author{Yangyi Cao$^{1}$ \hspace{0.65cm} Yuanhang Li$^{1}$ \hspace{0.65cm} Lan Chen$^{1}$ \hspace{0.65cm} Qi Mao$^{1,\textsuperscript{\Letter}}$ \\
$^1$MIPG, Communication University of China \\
$\{$caoyangyicuc, yuanhangli, qimao$\}$@cuc.edu.cn, chenlaneva@mails.cuc.edu.cn}

\maketitle
\let\thefootnote\relax\footnotetext{\textsuperscript{\Letter} Corresponding Author}

\begin{abstract}
We propose MLV-Edit, a training-free, flow-based framework that address the unique challenges of minute-level video editing. 
While existing techniques excel in short-form video manipulation, scaling them to long-duration videos remains challenging due to prohibitive computational overhead and the difficulty of maintaining global temporal consistency across thousands of frames.
To address this, MLV-Edit employs a divide-and-conquer strategy for segment-wise editing, facilitated by two core modules: 
Velocity Blend rectifies motion inconsistencies at segment boundaries by aligning the flow fields of adjacent chunks, eliminating flickering and boundary artifacts commonly observed in fragmented video processing; and Attention Sink anchors local segment features to global reference frames, effectively suppressing cumulative structural drift.
Extensive quantitative and qualitative experiments demonstrate that MLV-Edit consistently outperforms state-of-the-art methods in terms of temporal stability and semantic fidelity.
\end{abstract}


\section{Introduction}
\label{sec:intro}

Text-guided video editing\cite{wu2023tune,qi2023fatezero} aims to precisely manipulate the content of a source video according to textual prompts, while faithfully preserving the appearance and content of the unedited regions. 
In recent years, text-guided video editing makes significant progress, driven by the rapid development of powerful diffusion models\cite{ho2020denoising, li2024starvid, chen2025univid, wan2025wan}. 
Nevertheless, most state-of-the-art video editing methods are primarily designed for and evaluated on short video clips, typically lasting only a few seconds.
Directly extending these methods to long-form or minute-long videos remains challenging, due to prohibitive computational costs and the difficulty of maintaining consistent temporal coherence across extended sequences.

To ensure temporal consistency in editing results, several methods employ inversion-based\cite{song2020denoising} techniques to preserve visual fidelity\cite{tokenflow2023,kara2024rave,zhang2025adaflowefficientlongvideo}; Although effective for short clips, these approaches exhibit rapidly increasing memory consumption and computational cost as video duration grows, rendering them impractical for long-form videos.
Meanwhile, recent efficient architectures such as Diffusion Transformers (DiT)\cite{ jiang2025vace, li2025ic, li2025five} are fundamentally limited by their reliance on fixed-length context tokens or restricted attention windows, making them unsuitable for long videos.
Together, these architectural and methodological limitations underscore the urgent need for a scalable and duration-independent video editing framework.

\begin{figure}[t]
\centering
\includegraphics[width=1\linewidth]{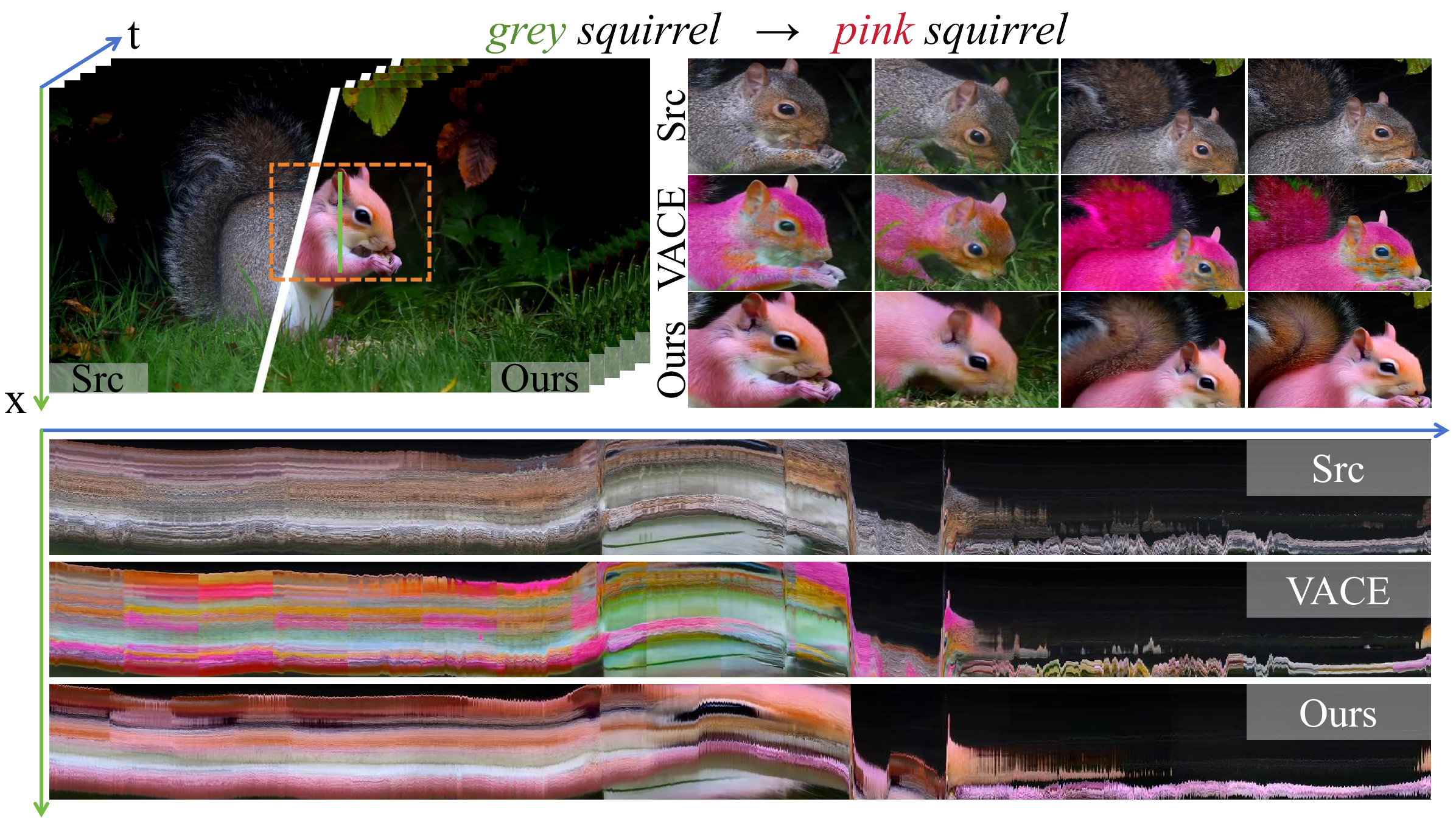}
\vspace{-5 mm}
\caption{\textbf{Temporal inconsistencies in long video editing}. Temporal slices are visualized by extracting pixels along a fixed vertical line across frames. Existing segmented editing methods fail to preserve consistent editing effects between segments. By employing Velocity Blend and Attention Sink, MLV-Edit effectively maintains coherent temporal evolution over long video sequences. }
\label{fig:Super-resolution}
\vspace{-5 mm}
\end{figure}

An intuitive and effective approach to bypass these computational and architectural scalability limitations is to adopt a divide-and-conquer paradigm, utilizing established short-video editing methods\cite{kara2024rave,tokenflow2023,jiang2025vace,li2025five} to process long videos segment by segment.
However, naive segment-wise editing and direct segment splicing fundamentally fail to maintain editing quality, leading to severe temporal inconsistencies throughout the long video, as shown in \figref{Super-resolution}.
This problem manifests primarily in two aspects: a) \textbf{Boundary Discontinuity}: Without explicit cross-segment coherence control, noticeable flickering or jitter emerges at segment boundaries.
b) \textbf{Effect Drift}: Independently editing each segment often produces inconsistent editing effects across adjacent segments, thereby undermining global coherence and editing fidelity.
Methods such as AdaFlow\cite{zhang2025adaflowefficientlongvideo} attempt to mitigate these issues by introducing explicit cross-segment blending mechanisms based on keyframes.
However, they still fail to fully resolve fundamental problems, including boundary discontinuity and effect drift, particularly under long-range temporal dependencies and strict editing uniformity requirements.

\begin{figure*}[t]
\centering
\includegraphics[width=1\linewidth]{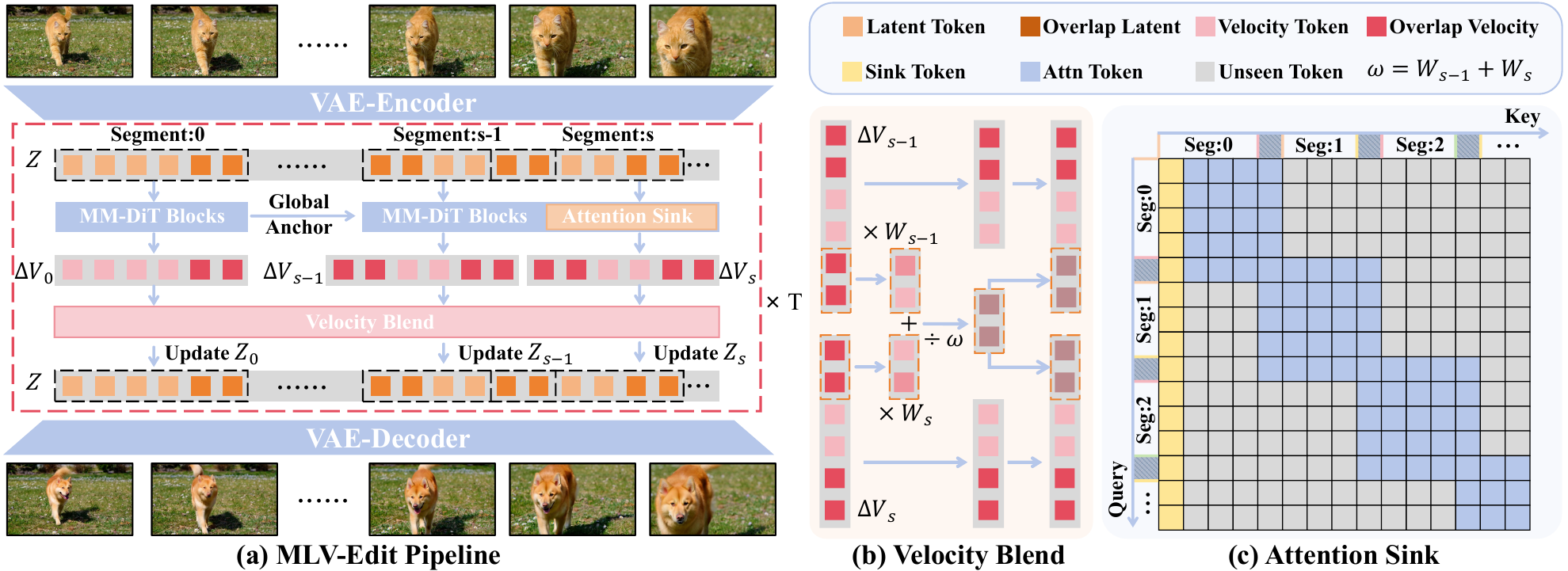}
\vspace{-5 mm}
\caption{\textbf{The framework of the proposed MLV-Edit.} (a) The overall pipeline of MLV-Edit for long video editing. MLV-Edit first encodes the source video into the latent space and partitions it into multiple overlapping segments. (b) Velocity Blend fuses the velocity fields in overlapping regions between adjacent segments. (c) Attention Sink caches the key and value pairs from the first frame and injects them into subsequent segments. }
\label{fig:Framework}
\vspace{-5 mm}
\end{figure*}

To this end, we propose \textbf{MLV-Edit}, a training-free, scalable, and length-independent framework for long video editing.
This framework eliminates the inherent length constraints of existing models without introducing significant computational overhead, enabling consistent and high-quality editing for video sequences of arbitrary duration.
To account for boundary discrepancies between adjacent segments, MLV-Edit adopts an overlapping segmentation strategy, in which neighboring segments intentionally overlap to capture transitional dynamics. 
Within these overlapped regions, MLV-Edit introduces \textbf{Velocity Blend} to mitigate splicing inconsistency and ensure smooth temporal transitions. 
Furthermore, MLV-Edit employs \textbf{Attention Sink} to maintain global semantic coherence throughout the video and effectively prevent editing effect drift.

To comprehensively evaluate the effectiveness of MLV-Edit, we construct MLV-EVAL, a minute-level benchmark that covers diverse long-video editing scenarios.
Extensive experiments on MLV-EVAL demonstrate that MLV-Edit achieves state-of-the-art performance in terms of temporal consistency and visual stability across long video sequences.
These results indicate that MLV-Edit provides a practical and robust solution for real-world long video editing applications.

Our main contributions are summarized as follows:
\begin{itemize}
    \item We present MLV-Edit, a novel training-free framework that enables consistent and high-quality editing for arbitrarily long video sequences. 
    
    \item We introduce Velocity Blend and Attention Sink to address boundary discontinuity and effect drift, thereby ensuring temporal consistency in long video editing results.

    \item We construct MLV-EVAL, a comprehensive minute-level benchmark for evaluating long video editing performance across diverse scenarios.
\end{itemize}

\section{Method}

Our proposed MLV-Edit builds upon the Wan-Edit framework~\cite{li2025five}, which is briefly reviewed in~\subsecref{Preliminary}.
Subsequently, we introduce the overall pipeline of MLV-Edit in~\subsecref{Overall Framework}.
We then detail our two key innovations: \textbf{Velocity Blend} (\subsecref{Velocity Blend}), designed to mitigate boundary discontinuities, and the \textbf{Attention Sink} mechanism (\subsecref{Attention Sink}), which ensures the long-term consistency of editing effects.

\subsection{Preliminary}
\label{subsec:Preliminary}
Wan-Edit~\cite{li2025five} approaches video editing by formulating the task as a continuous transport process between the source latent $Z_{t_i}^{\mathrm{src}}$ and the target latent $Z_{t_i}^{\mathrm{tar}}$. 
This process is driven by the velocity field difference $\Delta V_{t_i}$, defined as:
\begin{equation}
    \label{eq:delta_v}
    \Delta V_{t_i} = V(Z_{t_i}^{\mathrm{tar}}, t_i, P^{\mathrm{tar}}) - V(Z_{t_i}^{\mathrm{src}}, t_i, P^{\mathrm{src}}),
\end{equation}
where $t_i$ denotes the $i$-th sampling timestep, and $P^{\mathrm{src}}$ and $P^{\mathrm{tar}}$ represent the source and target textual descriptions, respectively.
The target latent is explicitly constructed as $Z^{\mathrm{tar}}_{t_i} = Z^{\mathrm{edit}}_{t_i} + Z^{\mathrm{src}}_{t_i} - X^{\mathrm{src}}$, where $X^{\mathrm{src}}$ denotes the initial source video latent. 
Consequently, the editing trajectory evolves according to the following sampling process:
\begin{equation}
    \label{eq: wanedit}
Z_{t_{i-1}}^{\mathrm{edit}} = Z_{t_i}^{\mathrm{edit}} + (t_{i-1} - t_i) \Delta V_{t_i},
\end{equation}
where $\Delta V_{t_i}$ captures the semantic variation required to guide the editing process.
For simplicity, we omit the timestep subscript $t_i$ in the following discussion.

\subsection{Overall Framework}
\label{subsec:Overall Framework}
As illustrated in \figref{Framework}(a), given a source video, MLV-Edit initially leverages a pre-trained VAE~\cite{wan2025wan} to encode the entire video into a latent sequence $X^{\mathrm{src}}$.
This sequence is partitioned into $m$ segments of equal length with a temporal overlap of $k$ frames, providing a shared temporal buffer for smooth transitions.
Subsequently, each segment undergoes editing via Wan-Edit~\cite{li2025five}.
To mitigate boundary flickering, the \textbf{Velocity Blend} module smooths the transition between adjacent segments by blending the velocity field difference within the overlapping frames.
During segment-wise editing, \textbf{Attention Sink} establishes a global semantic anchor by propagating key and value pairs of the initial latent to all subsequent segments, thereby preserving semantic consistency across the entire video.
The detailed algorithm is presented in \algref{mlv-edit}.

\begin{algorithm}[t]
\setstretch{0.89}
\caption{MLV-Edit Step}
\label{mlv-edit}
\KwIn{edited latent $Z^{\text{edit}}$,
      source video latent $X^{\text{src}}$,
      timestep $t_i$,
      source prompt $P^{\text{src}}$,
      target prompt $P^{\text{tar}}$, 
      number of segments $m$,
      Velocity Blend formula $F_{VB}$}
\KwOut{updated latent $Z^{\text{edit}}$}

Sample noise $N_{t_i} \sim \mathcal{N}(0, I)$ 

$Z^{\text{src}} \gets (1 - t_i)\, X^{\text{src}} + t_i\, N_{t_i}$

$Z^{\text{tar}} \gets Z^{\text{edit}} + Z^{\text{src}} - X^{\text{src}}$

\For{$s = 0,1,...,m$}{
\fbox {\parbox{0.78\linewidth}{

  \textit{During} $V^{\text{tar}}_{s} \gets V(Z^{\text{tar}}_{s}, t_i, P^{\text{tar}})$
  
  \If{$s = 0$}
  {
    cache $K_{0}^{0}, V_{0}^{0}$
  }
  \Else{
    $K_{s} = [K_{0}^{0},\, K_{s}^{(0:n)}]$ 
    
    $V_{s} =[V_{0}^{0},\, V_{s}^{(0:n)}]$ 
    
    $\text{Attention}\big(Q_s,K_s,V_s \big)$}
  }}
  \tcp{\textcolor{red}{\textbf{Attention Sink}}}
  
  \vspace{0.3ex}

  $V^{\text{src}}_{s} \gets V(Z^{\text{src}}_{s}, t_i, P^{\text{src}})$

  $\Delta V_{s} \gets V^{\text{tar}}_{s} - V^{\text{src}}_{s}$

\fbox {\parbox{0.78\linewidth}{
  \If{$s > 0$}{
    $\Delta V_{s}  \gets F_{\text{VB}}\big(\Delta V_{s-1},\, \Delta V_{s}\big)$ 
  }
  }}
  \tcp{\textcolor{red}{\textbf{Velocity Blend}}}
}

$Z^{\text{edit}} \gets Z^{\text{edit}} + (t_{i-1} - t_i)\, \Delta V$

\textbf{return} $Z^{\text{edit}}$
\end{algorithm}

\subsection{Velocity Blend}
\label{subsec:Velocity Blend}

The naive ``segment-and-stitch'' strategy treats segments in isolation, causing \textit{boundary discontinuity}.
When employing Wan-Edit~\cite{li2025five} as the pipeline, this independent processing induces abrupt changes in $\Delta V$ (defined in~\eqnref{delta_v}) at segment boundaries.
Such discontinuities in $\Delta V$ lead to significant divergence in the editing direction, manifesting as visual artifacts like flickering.
To resolve this, we introduce \textbf{Velocity Blend}, which harmonizes $\Delta V$ within overlapping regions.
Specifically, for an overlapping region of $k$ frames, we compute the blended velocity field $\Delta \tilde{V}$ by performing a weighted average of the tail of the current segment ($s-1$) and the head of the next segment ($s$):
\begin{equation}
\scalebox{1.23}{\(\Delta \tilde{V}_j =\frac{ W(n-k+j) \cdot \Delta V_{s-1}(n-k+j)+W(j) \cdot \Delta V_{s}(j)}{W(n-k+j) + W(j)}\)},
\label{eq:velocity_blend_v}
\end{equation}
where $ j\in[0,k]$, $\Delta V_{s-1}$ and $\Delta V_{s}$ refer to the velocity fields of the adjacent segments, and $n$ is the total segment length.
The position-dependent weight $W(\cdot)$ is defined as a symmetric triangular window to prioritize central frames:
\begin{equation}
W(\tau) = 1 - \left| 1 - \frac{2(\tau + 0.5)}{n} \right|,
\label{eq:velocity_blend_w}
\end{equation}
where $\tau$ represents the temporal index within a segment.
By explicitly blending the velocity field difference in the transitional buffers, we ensure a consistent semantic guidance across segments, resulting in smooth editing.

\subsection{Attention Sink}
\label{subsec:Attention Sink}
While Velocity Blend effectively mitigates discontinuities at segment boundaries, it is insufficient to guarantee long-term consistency throughout the entire video duration.
Specifically, due to the stochastic nature of diffusion models, simply using identical prompts does not guarantee identical editing effects across different segments.
Consequently, these inter-segment variations manifest as \textit{effect drift}, where visual attributes—such as subject identity, texture, or structure—fluctuate or diverge significantly across the video timeline.

To address this challenge, we introduce the \textbf{Attention Sink} mechanism, which establishes a stationary global semantic anchor to provide an explicit reference for all subsequent segments.
Since the VAE~\cite{wan2025wan} encodes the first frame independently without temporal downsampling, it retains the highest semantic fidelity. 
Consequently, we adopt this frame as the global anchor.
Specifically, for the initial segment, the self-attention follows the standard formulation:
\begin{equation}
\text{Attention}(Q,K,V) = \text{Softmax}\left(\frac{QK^T}{\sqrt{d}}\right)V,
\label{eq:self-attn}
\end{equation}
where we cache the Key ($K_0^0$) and Value ($V_0^0$) pairs of the first frame.
For all subsequent segments, we prepend this cached global anchor to the beginning of the current segment's key and value matrices:
\begin{equation}
K_{s} = [K_{0}^{0},\, K_{s}^{(0:n)}], \quad
V_{s} =[V_{0}^{0},\, V_{s}^{(0:n)}],
\label{eq:self-attn-concat}
\end{equation}
where the superscript indicates the temporal index within a segment.
Consequently, the attention computation is updated to:
\begin{equation}
\text{Attention}(Q_{s},K_{s},V_{s}) = \text{Softmax}(\frac{Q_{s} K_{s}^T}{\sqrt{d}}) V_{s}.
\label{eq:self-attn-update}
\end{equation}

By enforcing alignment with the global anchor at every denoising timestep, the Attention Sink strategy effectively suppresses semantic drift, ensuring consistent editing effects throughout long videos.

\begin{table*}[!t]
\centering
\vspace{-3 mm}
\caption{\textbf{Quantitative Evaluation. }
The best and second-best values are highlighted in \colorbox{pearDark!20}{blue} and \colorbox{mycolor_green}{green}, respectively.
}
\resizebox{\linewidth}{!}{%
\begin{tabular}{ccccccccccc}
\hline
\multicolumn{2}{c}{} & \multicolumn{1}{c}{\textbf{Subject Consistentcy}}  & \multicolumn{2}{c}{\textbf{Semantic Consistentcy}}  & \multicolumn{5}{c}{\textbf{Temporal Consistency}} & \multicolumn{1}{c}{\textbf{Fidelity}}\\ \cmidrule(r){3-3} \cmidrule(r){4-5} \cmidrule(r){6-10} \cmidrule(r){11-11}   
\multicolumn{2}{c}{\multirow{-2}{*}{\textbf{Method/Metrics}}} & \textbf{DINO} \textbf{($\uparrow$)} & \textbf{CLIP-T} \textbf{($\uparrow$)} & \textbf{ViCLIP-T} \textbf{($\uparrow$)} & \textbf{DOVER} \textbf{($\uparrow$)} & \textbf{CLIP-F} \textbf{($\uparrow$)} & \textbf{FS.clip} \textbf{($\uparrow$)}  & \textbf{Warp-Err}$_{(\times 10^{-4})}$ \textbf{($\downarrow$)} & \textbf{Seg.warperr}$_{(\times 10^{-4})}$ \textbf{($\downarrow$)} & \textbf{M.PSNR} \textbf{($\uparrow$)}  \\ 
\midrule
& RAVE\cite{kara2024rave}  &  0.977 &  25.29 &   \colorbox{mycolor_green}{26.92} &  0.809  & 0.972 & 0.887 & 28.54  &  27.45 &  17.42 \\
& VACE\cite{jiang2025vace}  &  \colorbox{mycolor_green}{0.991} &  23.09 &   22.38 &  0.860  & \colorbox{mycolor_green}{0.992} & 0.928 & \colorbox{mycolor_green}{5.446} &  \colorbox{mycolor_green}{5.312} &  \colorbox{mycolor_green}{26.15} \\
&  AdaFlow\cite{zhang2025adaflowefficientlongvideo} &  0.982 &  25.97 &   24.99 &  0.876  &  0.986 & \colorbox{mycolor_green}{0.942} & 6.294 & 6.216 &  21.76  \\
&  TokenFlow\cite{tokenflow2023}  &  0.986 &  \colorbox{mycolor_green}{27.36} &   26.70 &  \colorbox{mycolor_green}{0.879}  & 0.988 & 0.938  & 8.522 &  8.623 &  22.45 \\
& VideoPainter\cite{bian2025videopainter}  &  0.974 &  23.68 &   24.43 &  0.849  & 0.982 & 0.875 & 19.99 &  21.04 &  19.27 \\
\multicolumn{1}{c}{} & \textbf{Ours} &  \colorbox{pearDark!20}{0.992} & \colorbox{pearDark!20}{27.48} &   \colorbox{pearDark!20}{27.77} &  \colorbox{pearDark!20}{0.883}  & \colorbox{pearDark!20}{0.994} & \colorbox{pearDark!20}{0.946} &   \colorbox{pearDark!20}{5.254} & \colorbox{pearDark!20}{5.192}  &  \colorbox{pearDark!20}{30.33}  \\
\bottomrule
\end{tabular}%
}
\label{tab:tab_baseline}
\vspace{-3 mm}
\end{table*}

\section{Experiments}

\begin{figure*}[t]
\centering
\includegraphics[width=1\linewidth]{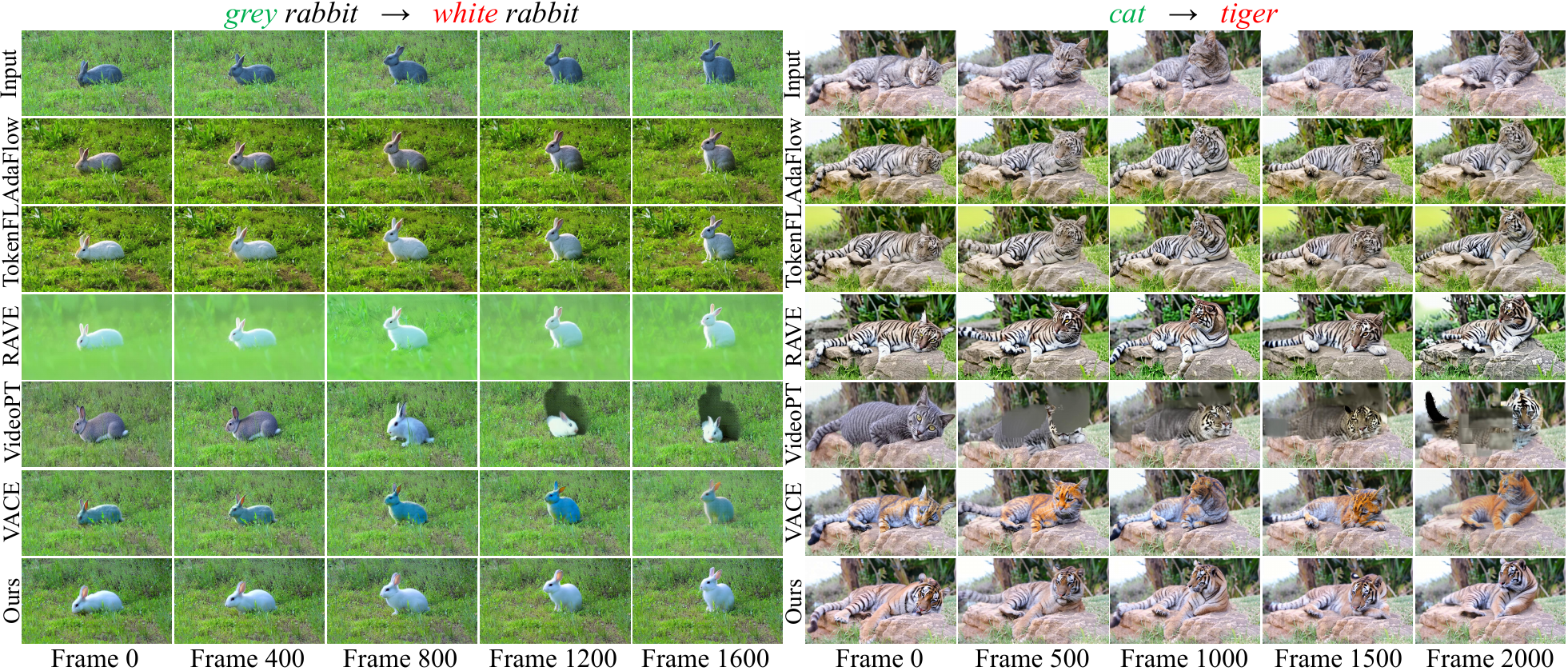}
\caption{\textbf{Comparisons of MLV-Edit with video editing baselines}. The MLV-Edit not only effectively suppresses the flickering and artifacts at the segment boundaries, but also preserves consistent editing effects throughout the long video sequences.}
\label{fig:qualities}
\vspace{-3 mm}
\end{figure*}

\subsection{Experimental Settings \label{Experimental Settings}}

\Paragraph{Implementation Details.} 
We implement our framework using the Wan2.1-T2V-1.3B model~\cite{wan2025wan} as the backbone.
We set the total number of timesteps $T$ to $25$ and the classifier-free guidance (CFG) scale to $7.5$.
For the temporal segmentation settings, we configure the segment length as $n=21$ and the overlap length as $k=5$.
All input videos are standardized to a resolution of $832 \times 480$.
Experiments are conducted on a single NVIDIA A800 GPU.

\Paragraph{Benchmark.} 
To comprehensively evaluate performance on long video editing tasks, we construct a new minute-level benchmark named \textbf{MLV-EVAL}.
This benchmark comprises $75$ videos collected from public online sources and the existing LongV-EVAL dataset~\cite{zhang2025adaflowefficientlongvideo}.
Each video spans a duration of $1$ to $2$ minutes, covering diverse content categories such as humans, animals, and plants.
For each sample, we utilize GPT-4~\cite{achiam2023gpt} to generate a source caption along with two high-quality editing prompts, including various tasks such as attribute modification and category replacement.

\Paragraph{Evaluation Metrics.} 
We quantitatively assess our method across four key dimensions:
\begin{compactitem}
\item \textbf{Subject Consistency.}
We utilize DINO\cite{caron2021emerging} to assess structural consistency of the edited subjects across frames.
\item \textbf{Semantic Consistency.}
We employ CLIP-T~\cite{radford2021learning} and ViCLIP-T~\cite{wang2023internvid} to quantify the semantic alignment between the output video and the target text prompt at both the frame-level and video-level.
\item \textbf{Temporal Consistency.}
We utilize DOVER~\cite{wu2023exploring}, CLIP-F~\cite{radford2021learning}, and Warp-Err \cite{lai2018learning} to evaluate the temporal consistency of the edited results. 
Specifically, DOVER~\cite{wu2023exploring} measures global temporal artifacts, while CLIP-F~\cite{radford2021learning} assesses the similarity between adjacent frames. Warp-Err~\cite{lai2018learning} quantifies temporal smoothness by calculating the optical flow difference between the original and edited videos. 
Furthermore, to evaluate global consistency and inter-segment smoothness, we construct \textit{Frame-Skip CLIP-F} (\textit{FS.clip}) to compute the similarity between frames across different segments. 
We also introduce \textit{Segment Warp-Err} (\textit{Seg.warperr}) to quantify the transition smoothness at the segment boundaries.
\item \textbf{Fidelity.} To evaluate the fidelity of the unedited regions, we employ Masked PSNR (M.PSNR) \cite{huynh2008scope} to measure the pixel-level reconstruction quality within these areas.
\end{compactitem}

\Paragraph{Compared Methods.}
To verify the effectiveness of MLV-Edit, we compare it with 
inversion-based approaches (TokenFlow \cite{tokenflow2023} (TokenFL), RAVE \cite{kara2024rave}, and AdaFlow \cite{zhang2025adaflowefficientlongvideo}) and DiT-based inversion-free approaches (VACE \cite{jiang2025vace} and VideoPainter \cite{bian2025videopainter}). 
Among these, AdaFlow\cite{zhang2025adaflowefficientlongvideo} and VideoPainter\cite{bian2025videopainter} (VideoPT) are specifically designed for long video editing. 
For the remaining baselines tailored for short videos, we adapt them to the long-video setting using a naive ``segment-and-stitch'' strategy, where partitioned clips are processed sequentially.

\begin{table*}[!t]
\centering
\vspace{-3 mm}
\caption{\textbf{Quantitative Comparison of  Ablation Study.}
The best and second-best values are highlighted in \colorbox{pearDark!20}{blue} and \colorbox{mycolor_green}{green}, respectively.
}
\resizebox{\linewidth}{!}{%
\begin{tabular}{ccccccccccc}
\hline
\multicolumn{2}{c}{} & \multicolumn{1}{c}{\textbf{Subject Consistentcy}}  & \multicolumn{2}{c}{\textbf{Semantic Consistentcy}}  & \multicolumn{5}{c}{\textbf{Temporal Consistency}} & \multicolumn{1}{c}{\textbf{Fidelity}}\\ \cmidrule(r){3-3} \cmidrule(r){4-5} \cmidrule(r){6-10} \cmidrule(r){11-11}   
\multicolumn{2}{c}{\multirow{-2}{*}{\textbf{Method/Metrics}}} & \textbf{DINO} \textbf{($\uparrow$)} & \textbf{CLIP-T} \textbf{($\uparrow$)} & \textbf{ViCLIP-T} \textbf{($\uparrow$)} & \textbf{DOVER} \textbf{($\uparrow$)} & \textbf{CLIP-F} \textbf{($\uparrow$)} & \textbf{FS.clip} \textbf{($\uparrow$)}  & \textbf{Warp-Err}$_{(\times 10^{-4})}$ \textbf{($\downarrow$)} & \textbf{Seg.warperr}$_{(\times 10^{-4})}$ \textbf{($\downarrow$)} & \textbf{M.PSNR} \textbf{($\uparrow$)}  \\ 
\midrule
& $F_{adj}=1$  &  \colorbox{mycolor_green}{0.991} &  \colorbox{mycolor_green}{26.26} &   26.04 &  \colorbox{mycolor_green}{0.882}  & \colorbox{mycolor_green}{0.993} & \colorbox{mycolor_green}{0.933} & 6.078 &  5.998 &   28.32 \\
& $F_{adj}=10$  &  0.983 &  25.95 &   25.86 &  0.856  & 0.985 &  0.904 & 9.758 &  9.379 &   25.67 \\
& $F_{fir}=10$  &  0.987 &  25.94 &   25.71 &  0.871  & 0.989 & 0.925 & 7.662 & 7.354 &  27.43 \\
& w/o Attention Sink   &  0.963 &  26.13 &   \colorbox{mycolor_green}{26.08} &  0.878  & 0.983 & 0.923 &  \colorbox{mycolor_green}{5.811} &  \colorbox{mycolor_green}{5.920} &   \colorbox{mycolor_green}{28.86} \\
\midrule
& $K=1$  &  0.989 &  26.02 &   25.89 &  0.880  & 0.992 &  0.919 & 6.182 &  6.088 &  27.51 \\
& $K=10$  &  \colorbox{mycolor_green}{0.991} &  \colorbox{mycolor_green}{26.10} &   25.88 &  \colorbox{mycolor_green}{0.882}  & \colorbox{mycolor_green}{0.993} &  \colorbox{mycolor_green}{0.928} &  \colorbox{mycolor_green}{5.679} &  \colorbox{mycolor_green}{5.605} &    \colorbox{mycolor_green}{29.12} \\
& w/o Velocity Blend  &  0.990 &  26.06 &   \colorbox{mycolor_green}{25.95} &  0.881  & 0.992 &  0.921 & 5.716 &  5.674 &   28.63 \\
\midrule
\multicolumn{1}{c}{} & \textbf{Ours} &  \colorbox{pearDark!20}{0.992} & \colorbox{pearDark!20}{27.48} &   \colorbox{pearDark!20}{27.77} &  \colorbox{pearDark!20}{0.883}  & \colorbox{pearDark!20}{0.994} &  \colorbox{pearDark!20}{0.946} &  \colorbox{pearDark!20}{5.254} &  \colorbox{pearDark!20}{5.192} &   \colorbox{pearDark!20}{30.33}  \\
\bottomrule
\end{tabular}%
}
\label{tab:tab_ablate}
\vspace{-5 mm}
\end{table*}

\subsection{Comparisons with Baselines}
\label{sec:Editing Performance Comparison}

\Paragraph{Quantitative Evaluation. }
\tabref{tab_baseline} presents the quantitative comparison between our MLV-Edit framework and the competing methods.
The results demonstrate that MLV-Edit consistently outperforms all evaluated baselines across four key dimensions: subject consistency, semantic consistency, temporal consistency, and fidelity, thereby validating its effectiveness in long video editing.
Most notably, MLV-Edit achieves a substantial margin of improvement in temporal consistency, significantly surpassing even those methods explicitly tailored for long videos, such as AdaFlow~\cite{zhang2025adaflowefficientlongvideo} and VideoPT~\cite{bian2025videopainter}.
The leading performance across these metrics validates that MLV-Edit effectively mitigates the critical challenges of stitching inconsistencies and content drift inherent in long-form video synthesis.

\Paragraph{Qualitative Evaluation. }
\figref{qualities} illustrates our qualitative comparison results with other baselines.
As the video length increases, most baseline methods exhibit noticeable effect drift. 
For example, TokenFlow\cite{tokenflow2023} and VideoPT\cite{bian2025videopainter} fail to maintain consistent attributes for the rabbit across distant frames.
Although AdaFlow~\cite{zhang2025adaflowefficientlongvideo} is explicitly designed for long video editing, it remains susceptible to inter-segment inconsistencies.
For instance, the tiger in the $1000$-th frame shows duplicated head regions.
Additionally, methods like RAVE~\cite{kara2024rave} tend to introduce unintended background modifications.
In stark contrast, our method consistently preserves global consistency across the entire sequence while maintaining high fidelity.

\begin{figure}[t]
\centering
\includegraphics[width=1\linewidth]{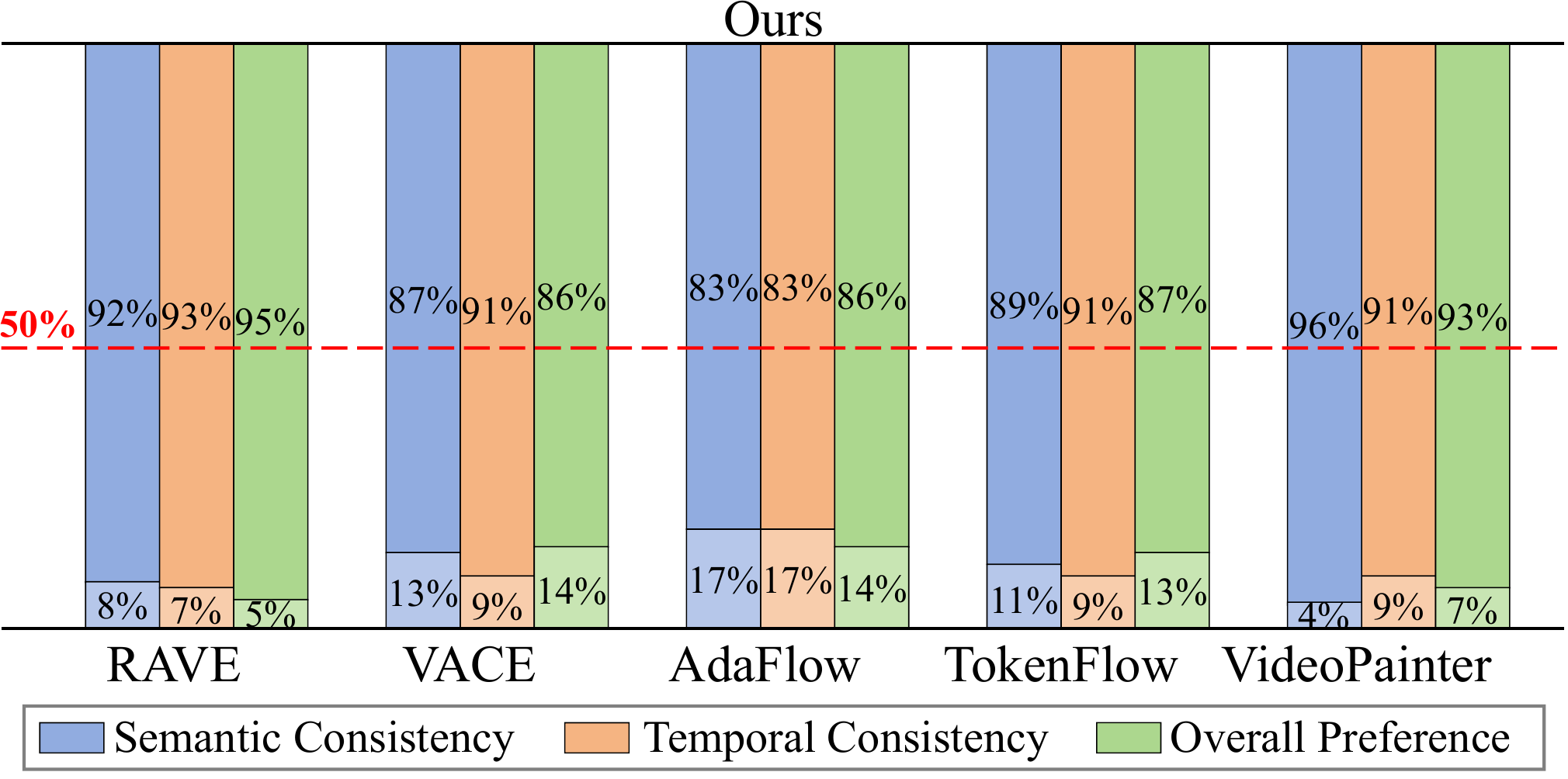}
\caption{\textbf{User study result}. The values presented reflect the proportion of users who favor our proposed method over comparative approaches.}
\label{fig:user-study}
\vspace{-5 mm}
\end{figure}

\subsection{User Study}

To further evaluate MLV-Edit, we conduct a user study using an A/B testing protocol. 
Participants are presented with a text prompt, the source video, and two edited results (ours vs. a baseline) in a randomized order to eliminate bias. 
For fairness, all videos are generated using identical prompts and random seeds. 
Participants then evaluate the results based on the following questions: 
\begin{compactitem}
\item  \textbf{Semantic Consistency}: Which video exhibits more consistent editing results?
\item \textbf{Temporal Consistency}: Which video presents smoother temporal transitions? 
\item  \textbf{Overall Preference}: Which video do you prefer?
\end{compactitem}
We recruit $20$ participants to evaluate a total of $30$ sets of editing results. 
As illustrated in \figref{user-study}, participants consistently favored MLV-Edit over the baselines across all dimensions, demonstrating its superior perceptual quality.
\begin{figure}[t]
\centering
\includegraphics[width=1\linewidth]{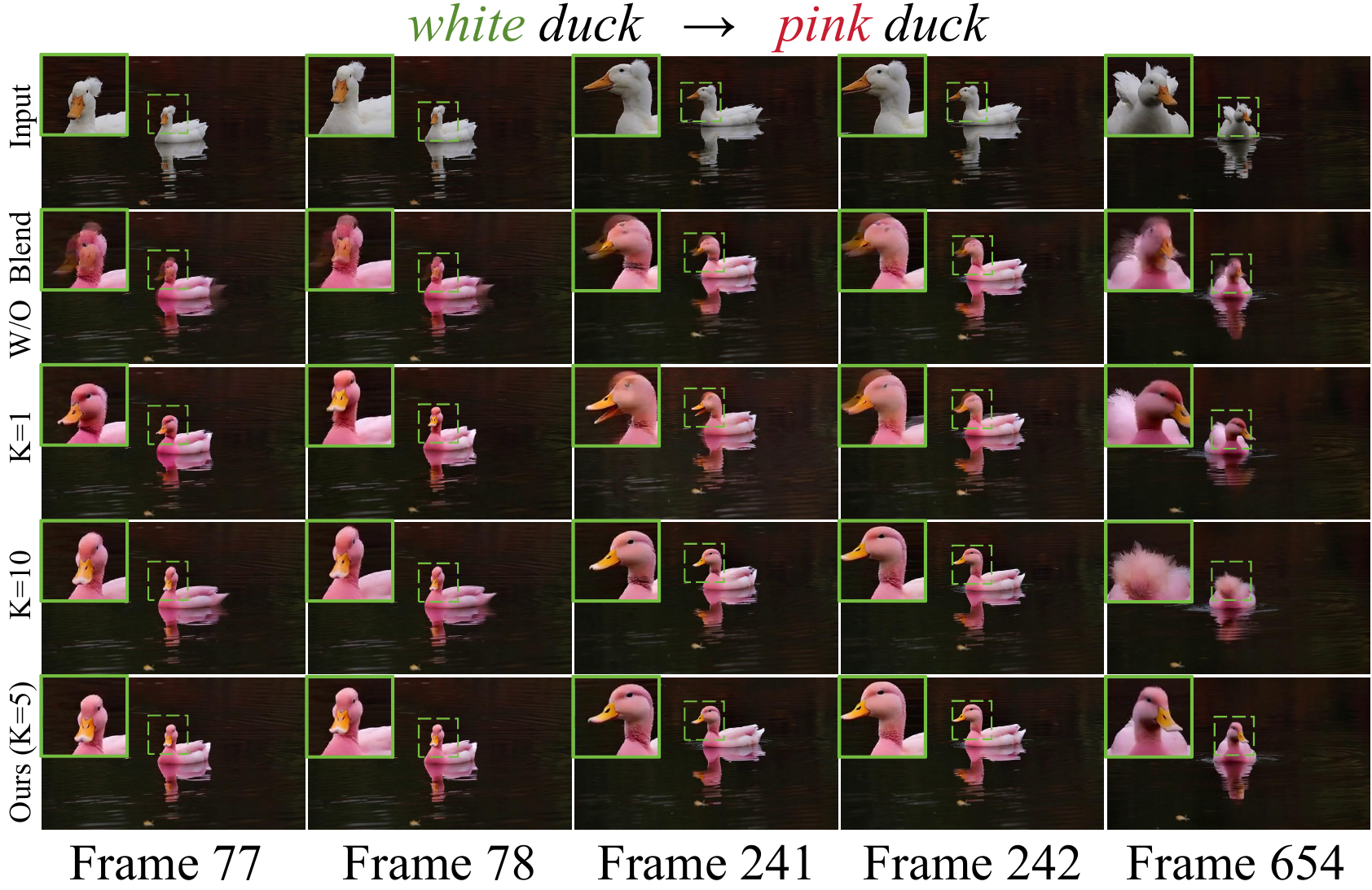}
\caption{\textbf{Ablation study on the influence of Velocity Blend}. }
\label{fig:ablate-blend}
\vspace{-5 mm}
\end{figure}
\subsection{Ablation Studies}

\Paragraph{Overlap length in Velocity Blend. }
To investigate the impact of the overlap length $k$, we evaluate our default setting ($k{=}5$) against three variants: disabling Velocity Blend, $k{=}1$, and $k{=}10$.
Qualitative results in \figref{ablate-blend} reveal that disabling Velocity Blend leads to severe transition artifacts, which cannot be adequately eliminated by a small overlap ($k{=}1$).
Conversely, an excessively large overlap ($k{=}10$) forces the blending of segments that may have significantly diverged, resulting in feature conflicts that lead to reconstruction failure.
In contrast, our default setting of $k{=}5$ effectively smooths transitions without introducing artifacts.
These observations are corroborated by the quantitative results in \tabref{tab_ablate}, where both insufficient and excessive overlaps lead to a decline in temporal consistency metrics, whereas our configuration yields the best performance especially \textit{Seg.warperr}.

\begin{figure}[t]
\centering
\includegraphics[width=1\linewidth]{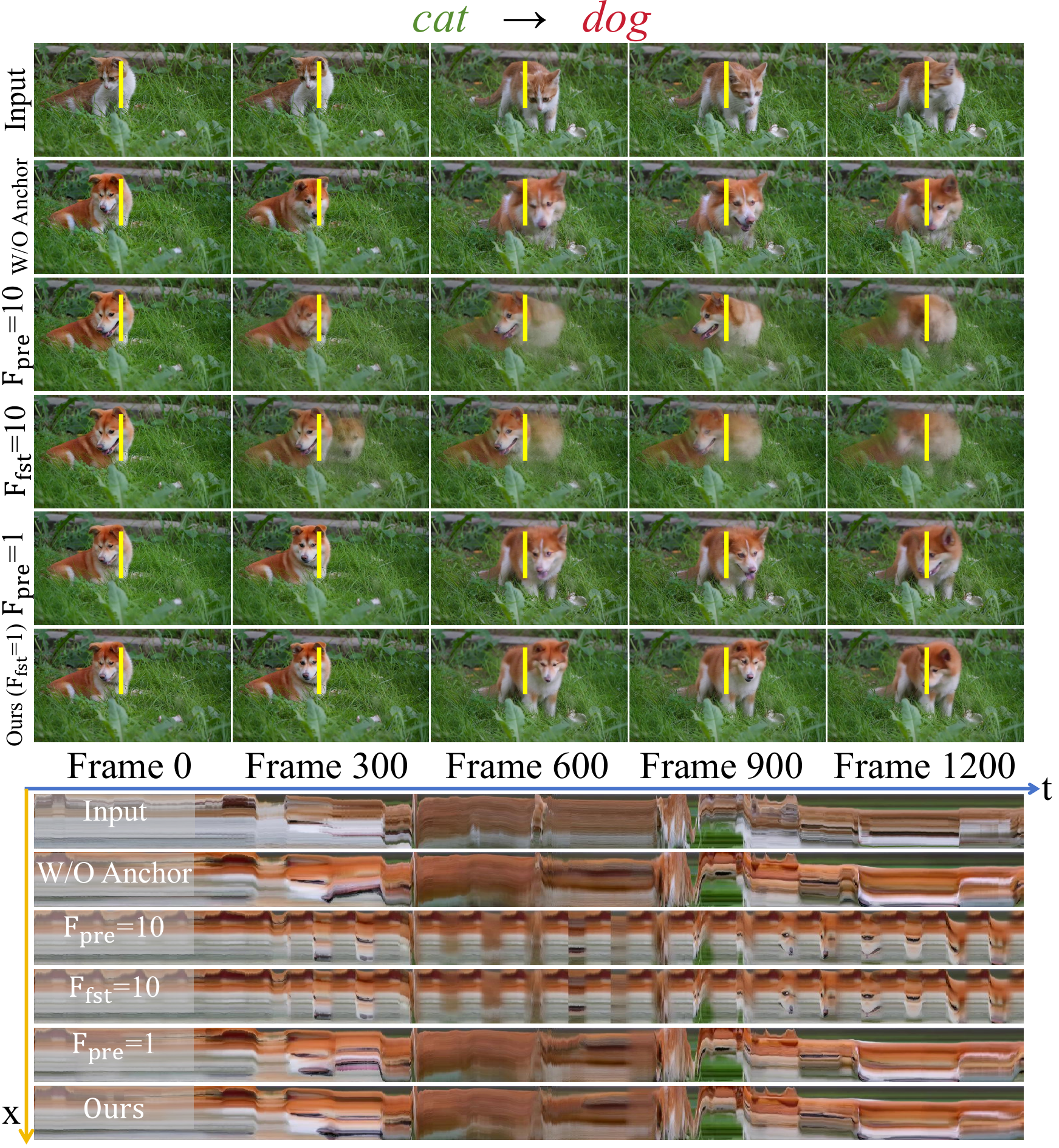}
\caption{\textbf{Ablation study on the effectiveness of Attention Sink}. For each frame, we visualize temporal slices by extracting pixels along a fixed vertical line over time. }
\label{fig:ablate-qkv}
\vspace{-5 mm}
\end{figure}

\Paragraph{Anchor selection for Attention Sink. } 
To validate the effectiveness of Attention Sink, we compare our approach against four alternatives: (1) using no anchor; (2)  using the first $10$ latents of the previous segment ($F_{\mathrm{pre}}{=}10$); (3) using the first $10$ latents of the initial segment ($F_{\mathrm{fst}}{=}10$); and (4) using the first single latent of the previous segment ($F_{\mathrm{pre}}{=}1$).
\figref{ablate-qkv} demonstrates that the absence of attention sink results in severe semantic drift.
Using multiple latents as anchors fails to mitigate this issue and instead introduces ghosting and blurring artifacts due to feature conflicts.
When reducing the anchor to a single latent, relying on the previous segment ($F_{\mathrm{pre}}{=}1$) still remains ineffective, yielding degraded consistency similar to the no-anchor setting.
In contrast, adopting the single initial latent($F_{\mathrm{fst}}{=}1$) effectively alleviates the effect drift, ensuring robust global consistency.
As validated quantitatively in~\tabref{tab_ablate}, our method secures the optimal performance in temporal metrics, confirming its robustness in maintaining long-term global consistency.

\section{Conclusion}

In this paper, we present MLV-Edit, a training-free and scalable framework for long video editing. By adopting a segment-wise editing strategy, MLV-Edit effectively extends short-video editing models to minute-level videos without additional training or length constraints.
To address key challenges in long video editing, including inter-segment inconsistencies and effect drift, we introduce Velocity Blend to smooth transitions across segments and Attention Sink to enforce global semantic consistency. These designs jointly enable coherent temporal evolution and stable semantic control over long video sequences.
Extensive experiments on the MLV-EVAL benchmark demonstrate that MLV-Edit consistently outperforms state-of-the-art methods in semantic alignment, temporal consistency, and visual stability. Overall, MLV-Edit provides an efficient and practical solution for high-quality long video editing and offers a promising direction for long-form video manipulation.

\bibliographystyle{MLV-Edit/IEEEbib.bst}
\bibliography{MLV-Edit/icme2025references.bib}

\end{document}